\pdfoutput=1

\documentclass[11pt]{article}

\usepackage[]{acl}

\usepackage{times}
\usepackage{latexsym}

\usepackage[T1]{fontenc}

\usepackage[utf8]{inputenc}

\usepackage{microtype}

\usepackage{graphicx}
\usepackage{subcaption}

\title{Simplifying Dataflow Dialogue Design}

\author{Joram Meron \\
  Telepathy Labs  GmBH, Zurich, Switzerland \\
  \texttt{joram.meron@telepathy.ai} }

\begin{document}
\maketitle
\begin{abstract}
In \citep{andreas2020task-oriented}, a dataflow (DF) based dialogue system was introduced, showing clear advantages compared to many commonly used current systems. This was accompanied by the release of SMCalFlow, a practically relevant, manually annotated dataset, more detailed and much larger than any comparable dialogue dataset.
Despite these remarkable contributions, the community has not shown further interest in this direction.
What are the reasons for this lack of interest? And how can the community be encouraged to engage in research in this direction?

One explanation may be the perception that this approach is too complex - both the the annotation and the system. 
This paper argues that this perception is wrong: 1) Suggestions for a simplified format for the annotation of the dataset are presented, 2) An  implementation of the DF execution engine is released\footnote{https://github.com/telepathylabsai/OpenDF}, which can serve as a sandbox allowing researchers to easily implement, and experiment with, new DF dialogue designs.
The hope is that these contributions will help engage more practitioners in exploring new ideas and designs for DF based dialogue systems.
 
\end{abstract}

\section{Introduction}

Traditional task oriented dialogue systems implement a pipeline consisting of NLU (natural language understanding, converting the user input to a structured representation), which feeds into a DM (dialogue manager), implementing dialogue state tracking (DST) as well as a policy to decide what action to take next. Both the NLU and DM components can take different forms.

Modern machine learning approaches have demonstrated large improvements in prediction accuracy for both the NLU and DM components (as well as end-to-end pipelines and other architectures), but for industrial applications, where strict control of the system behaviour is required, it is very common that the NLU is implemented as a machine learning based intent/entity classifier, while the DM is conceptually based on a finite state machine (FSM)-like paradigm, typically with the states and transitions manually defined.

The combination of intent/entity NLU, and FSM based DM offer controllability and reduces the training data requirements, but come at the cost of reduced expressiveness of user requests, as well as practical difficulty in scaling up the dialogue design as more interaction scenarios are added. 

\section{Dataflow Dialogues}

Microsoft's Semantic Machines (SM) introduced a dataflow dialogue system in \citep{andreas2020task-oriented}, which represents the user requests as rich compositional (hierarchical) expressions, which encode computational graphs. An engine executes these computations, which results in manipulating the computational graphs, generating an answer (possibly an error message), and optionally producing some side effects through API's to external services.

The prominent features of this system are:
\begin{itemize}
\item The system represents the dialogue history as a set of graphs, where each computational graph typically represents one user turn.
\item it has a {\it refer} operation to search over the current computational graphs (as well as external resources) which allows easy look-up and re-use of graph nodes which occurred previously in the dialogue.
\item it has a {\it revise} operation which allows modification and reuse of previous computations
\item it has an exception mechanism which allows convenient interaction with the user (e.g. asking for missing information, and resumption of the computation once the information is supplied).
\end{itemize}

These features correspond to essential phenomena in natural conversations (referring to previous turns, modifying previous requests, reacting to wrong information, etc.), which allows the system to handle these phenomena more effectively, and help mitigate the problem of DM scaling, compared to FSM-based DMs\footnote{No claim is being made here that the DF approach can {\it theoretically} represent more complex user requests than the FSM approach, rather that it can {\it practically} be significantly easier to handle complex requests using DF.}.

Extracting richer representation from the user requests, does increase the difficulty for the NLU task. To address this, the developers of the DF system released SMCalFlow - the largest dialogue dataset to date, manually annotated using the expressive compositional DF format, presumably the result of thousands of man hours of work. The dialogues in this dataset cover diverse common tasks, which can be relevant and interesting for developers of practical task oriented dialogue systems.

In the same paper, they also show that using the DF paradigm can be advantageous even without requiring manual preparation of rich annotations (demonstrated on the MultiWOZ dataset \citet{budzianowski-etal-2018-multiwoz}). Instead, the original intent/entity style annotation is automatically translated into DF expressions.

\section{Practical Applications}
Developers of actual applications, constrained by business goals, are often unable to take advantage of the recent advances achieved in dialogue system research. One of the most common reason for that is the need for the developers to be able to closely control the behaviour of the application, and be able to quickly adapt it whenever errors are discovered or specifications evolve.

For this reason, approaches allowing manual control (as opposed to black box approaches where control can only be asserted through engineering of the training data) are often preferred. 

One example is the use of FSM based DMs. These can be closely controlled, as they are based on manually created rules, which makes them very easy to use when creating simple applications. However, FSMs are notoriously difficult to scale up, especially when it comes to tasks with multiple possible execution paths.

The DF paradigm can be used for such applications, using manually crafted rules to closely control the application, while potentially reducing developer efforts as applications grow.

\section{Complexity}
More than a year has passed since the publication of the original DF paper, the release of SMCalFlow, and a related public leader-board. Despite the obvious theoretical and practical advantages of the DF approach, and the potential for further improvements, the community has shown only limited interest in following this work.

Why is this the case?
This paper hypothesizes that the reason may be a perception that the DF approach is inherently too complex (both the annotation, and the implementation), and that this perception is a result of two factors. First, the complexity of the annotation of SMCalFlow (which is the only dedicated DF dataset available), and second, the fact that the implementation of the SMCalFlow executable functions is excluded from the released code. The lack of documentation and of explanation of the design decisions behind the annotation, only exacerbates this perception.

In order to encourage the community to engage in this line of research, this paper tries to counter this misconception, by following two directions:
\begin{enumerate}
\item By releasing a basic implementation of the SMCalFlow execution engine.
Since this represents the work of one person during a limited time, only a part of the functionality of SMCalFlow is implemented, but it should allow readers to get familiar enough with the DF approach, so that they can easily implement their own ideas in this framework.
\item Explore (and demonstrate) ways to simplify SMCalFlow. Specifically, simplifying the annotation format in a way which reduces the effort for annotators (and readers), as well as for the  NLU component, while maintaining its ability to be correctly executed. 
\end{enumerate}

This paper also tries to emphasize that DF is not a monolithic system, with fixed functionality, applications, tools, formats and implementation. Rather, it is a general approach; a paradigm, which can have different interpretations and flavours, and which is ready for further research as well as practical uses.

\section{Simplifying SMCalFlow}

The DF approach is an object-oriented paradigm, which describes user requests as executable programs. Annotation is  practically a programming task, translating a user's request to a program, whose execution will satisfy the user's request. These programs consist of two types of functions: base DF-framework functions (like {\it refer} and {\it revise}), and application specific functions (e.g. {\it CreateEvent} for SMCalFlow).

In designing the application specific functions, the developers need to come up with a consistent design for a set of functions, which can be combined together to describe and execute user requests.

As mentioned above, multiple alternative designs are possible. Each design results in a trade-off between coverage of possible user requests, good programming practices (such as clarity, modularity and low complexity of the programs, and hence of the annotations), efficiency of execution, the demands it puts on the NLU component, etc.

In the following, a simplified annotation is presented, with the motivation to dispel the perception that SMCalFlow is too complex. The starting point of this work is SMCalFlow, since creating a new dataset was beyond the resources available for this work. The price to pay for this is twofold. First, the new design is tied to the original style. Creating a new dataset (or a completely new annotation for an existing dataset) would allow complete freedom in suggesting a new annotation style, but would require intensive labour. Second, once a new design has been defined for the annotation style, the original dataset needs to be transformed to the new design. The transformation process is a necessary step, but not the primary focus of this paper.

While DF is not inherently complicated, finding a good design is a challenging task. A novel aspect of this challenge is the need for the design to function correctly within the DF paradigm, e.g. use the {\it refer} and {\it revise} operators. Indeed, one of the motivations of this work is the hope that the community can suggest interesting new designs, which can serve as templates for further applications.

The work progressed by taking one original annotation at a time, defining the simplified annotation for it, and then implementing the needed DF functions (iteratively modifying the implementation to ensure new functions can interact consistently with previously implemented functions). Due to limited resources, only a small portion of the original expressions were inspected, concentrating on the domain of event creation/update, which covers only part of  SMCalFlow's more than 300 functions. Hopefully, the released implementation can serve as a useful example, and motivate interested researchers to expand this work.

Note that SM released a modified version of SMCalFlow \citep{platanios2021value-agnostic}, which also represents a kind of annotation simplification, but focusing on improving NLU accuracy, rather than understandability. This paper uses the original SMCalFlow version.

\subsection{Simplification Mechanism}
Due to the size of the dataset, manual simplification is clearly not feasible, therefore all simplifications were done programmatically. The full dataset was simplified, without trying to filter out some of the "irregular" annotations in the original dataset.
 
For convenience, the simplified format uses Python style expressions (as opposed to the Lisp style S-expressions in the original dataset), as this format is generally more familiar (the released system itself is written in Python).

The simplification was practically done by implementing a set of tree transformation rules, which convert specified sub-trees of the original expressions into simplified sub-trees (the transformation code is part of the release). A few tens of such rules were implemented
(additional rules will be needed as more SMCalFlow functions are implemented).

The simplification is applied to the whole dataset,
resulting in a simplified dataset, which can be used by the same pipeline used in the original paper.

\subsection{\label{simp-principles}Simplification Approach}

The design principles (which are heuristic, and not mutually exclusive) for the simplifications were: 1) Retain only necessary information in the annotation, 2) Avoid explicit logical steps, 3) Move logic from the annotation to the implementation, 4) Group and reuse repeating sequences of functions.

Practically this means: Try to omit any information which can be {\it deterministically} inferred from context (or possibly, which can be guessed). Keep only information which can not be inferred. Specifically, logical steps which can be inferred from context, are moved from the annotation into the implementation of the functions. For example, explicit type casts which are clear from the context can be omitted. Similarly, when needed information is missing in the user input, but can be inferred from the context, the simplified annotation should then include only the supplied information, and the logic to infer the missing information is left for the function implementation. 

Finally, the simplified annotation tries to avoid fragments of the original annotation which serve only "formal" purposes, and instead tries to style the annotation to be closer to a more natural/comprehensible description of the user requests. Please see the appendix for examples.

\subsection{Executing Simplified Annotations}

At execution time, an additional step (which can be viewed as the inverse of the dataset simplification step) translates the simplified annotation to a fully executable expression. This is done, again, by implementing tree transformation rules (for each function), which can add deterministically inferable missing information/steps (e.g. casting input to the right type, or performing other conversions/functions based on input type). 

\subsection{Simplification Results}

Since the original code to execute SMCalFlow was not released (and documentation not supplied), it is impossible to verify that the suggested simplifications in fact correctly implement the same logic (nor what that logic actually is) (in fact this was one of the motivations for this paper). Some examples of simplification are shown in the appendix, but it can only be left to the readers to inspect the simplified annotations and the code and draw their own conclusions.

\begin{table}
\centering
\begin{tabular}{ll}
\hline
  & Program Length \\
\hline
Original Annotation & (11, 37, 58) \\
Simplified Annotation & (2, 13, 22) \\
\hline
\end{tabular}
\caption{\label{program-length}
Program length of the two annotation styles. Length is measured as number of seq2seq target tokens, when translating user request to annotation. Showing (.25, .50, .75) quantiles over the entire dataset.}
\end{table}

Table \ref{program-length} shows the results of a comparison of the annotation lengths of the original and simplified annotations, confirming that the simplification does make the annotation significantly shorter (adding more simplification rules, as more functions are implemented, is expected to further reduce the length). In addition to being shorter, the simplified annotations are also more understandable (unfortunately, this can not be shown with simple objective measures), which should reduce annotation efforts when creating new training data.
 
To verify that the simplified annotations do not increase the burden on the NLU component, the translation pipeline of the original paper was used to train and evaluate seq2seq models using the simplified dataset. The result shows no degradation in translation accuracy (in fact a slight improvement in exact match, from 72.8\% to 73.8\%). With the significant decrease in annotation length, it could be expected that translation accuracy would actually improve. Possible  explanations why this did not happen could be that the simplifications are highly regular, or the fact that the coverage of the simplification rules is only partial.

\section{Further Work}

With the released execution code, deeper probing and exploration can become possible, compared to having access only to the SMCalFlow dataset. 

Particular areas of interest may include:

Evaluation: in addition to the exact-match metric for translation accuracy, other metrics can be used, such as comparison of execution results, graph structure similarity, etc. 

Using the graph structure: the graph structure (at different points of the execution) can be used by prediction models. For example, the user request could be recursively translated into a hierarchical graph using graph self attention as well as attention to previous graphs before/after execution.

Different design patterns which are beneficial to specific parts of the system. For example, the execution of a computation graph could emit various types of information which would then be useful for subsequent prediction models.

And of course, completing the design and implementation of a full working SMCalFlow system.

\section{Conclusion}
This paper argues that dataflow dialogue systems are worthy of more attention from the community.

In order to lower the barrier of entry into this field, a simplified version of a DF dataset, which is designed to be more easily understandable, has been described. This is accompanied by the release of a basic implementation of a DF system, which should allow interested researchers to easily implement their own designs and extensions. 

\newpage

\bibliography{df}

\cleardoublepage

\appendix

\section{Annotation Simplification Examples}
\label{sec:appendix}

In this section, two examples are given for annotation simplification. 

The annotations are shown in both the text and the computational graph formats. 
The graphs are constructed from the text, and are basically equivalent to the text format, but slight modifications occur during the graph construction process. 

Some details are given as to how the simplification principles mentioned in section \ref{simp-principles} are applied to the example annotations.

These examples are intended to demonstrate the feasibility of automatically simplifying the original annotation (further, or different, simplification is clearly possible). 

More important than the simplification process is the final form of the simplified annotations. The goal of this paper is to demonstrate that concise, understandable, and yet fully descriptive and executable annotations are possible, and that implementing them does not have to be complex.

The simplified form of the annotation can be used by the pipeline described in \citep{andreas2020task-oriented}, or by the methods described e.g. in \citep{platanios2021value-agnostic}, \citep{yin2021compositional}, \citep{mansimov2021semantic}.

\subsection{Example 1}
In this example, the user request is:

\textsf{ "Cancel my 10 AM tomorrow"}

\begin{figure}
\includegraphics[height=0.071\textheight,trim= 0.0in 0.5in 0.8in 0.5in]{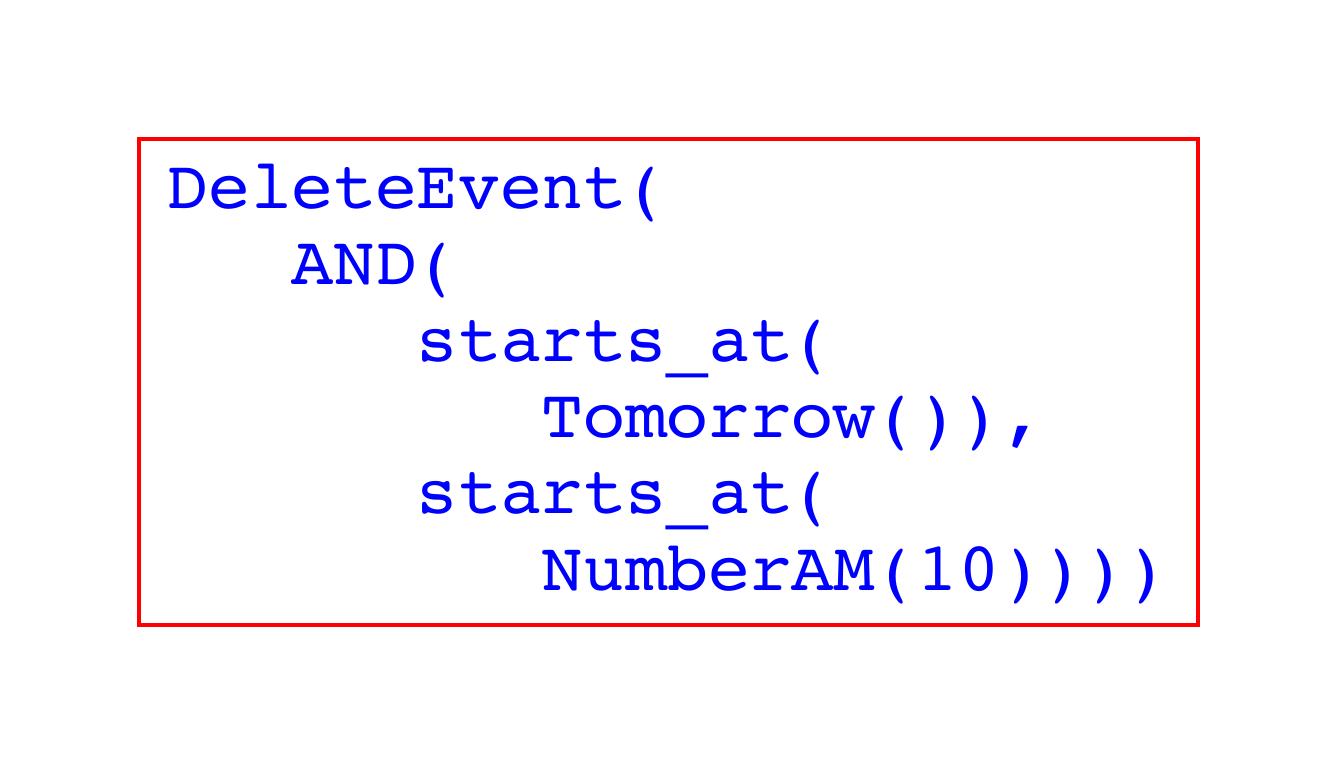}
\caption{\label{ex1-simp-txt} Example 1 - Simplified annotation}
\end{figure}

\begin{figure}
\includegraphics[height=0.4\textheight,trim= 0.8in 0.5in 0.8in 0.5in]{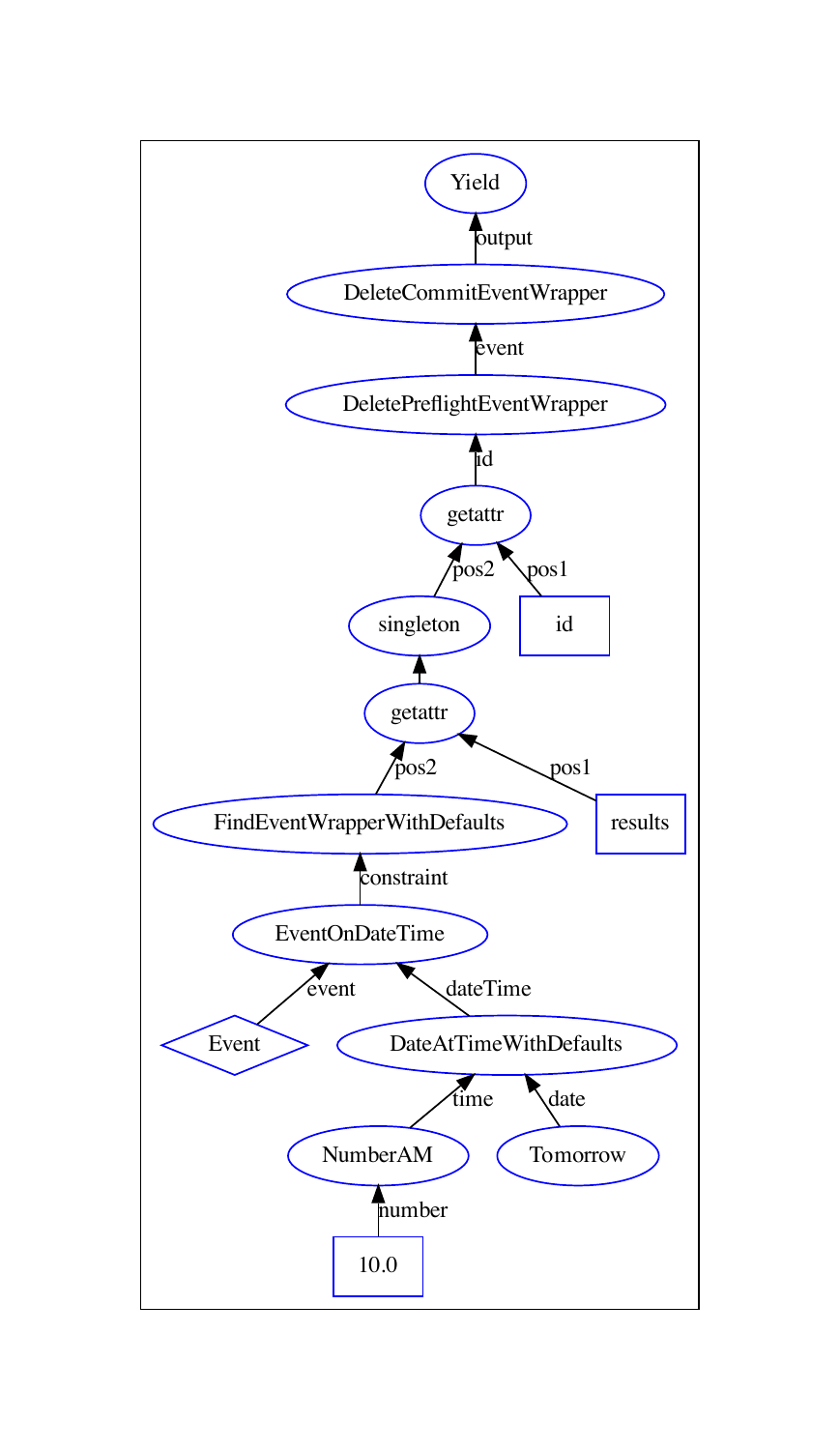}
\caption{\label{ex1-org} Example 1 - Original graph}
\end{figure}

\begin{figure}
\includegraphics[height=0.20\textheight,trim= 0.8in 0.5in 0.8in 0.5in]{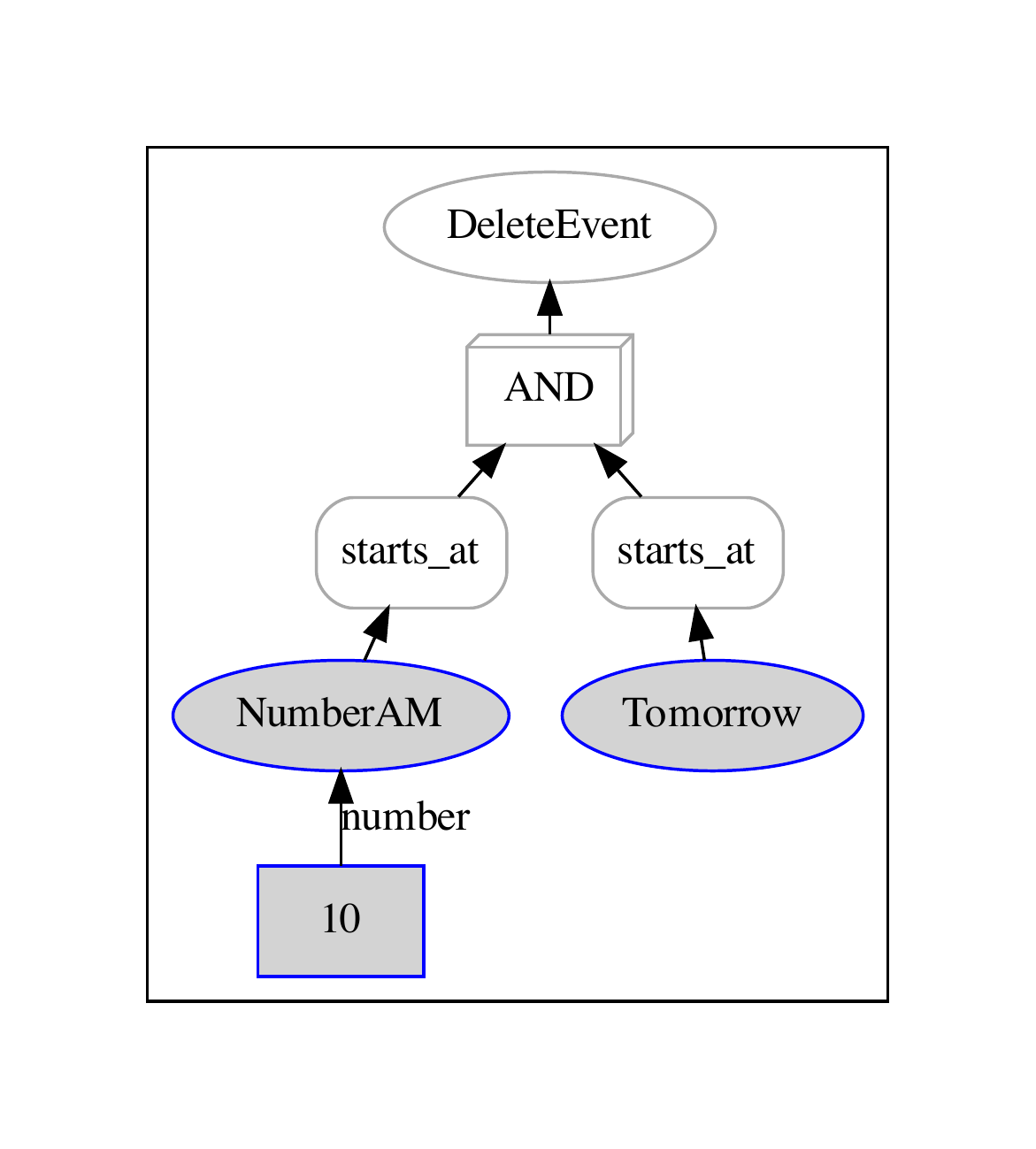}
\caption{\label{ex1-simp} Example 1 - Simplified graph}
\end{figure}

Figure \ref{ex1-org-txt} and \ref{ex1-org} show the original annotation for this request in text and graph forms, respectively, while figure \ref{ex1-simp-txt} and \ref{ex1-simp} show the same for the simplified annotation.

As explained in \ref{simp-principles}, simplification rules are used to remove operations which can be inferred from context. In this example:

\begin{figure}
\includegraphics[height=0.15\textheight,trim= 0.8in 0.5in 0.8in 0.5in]{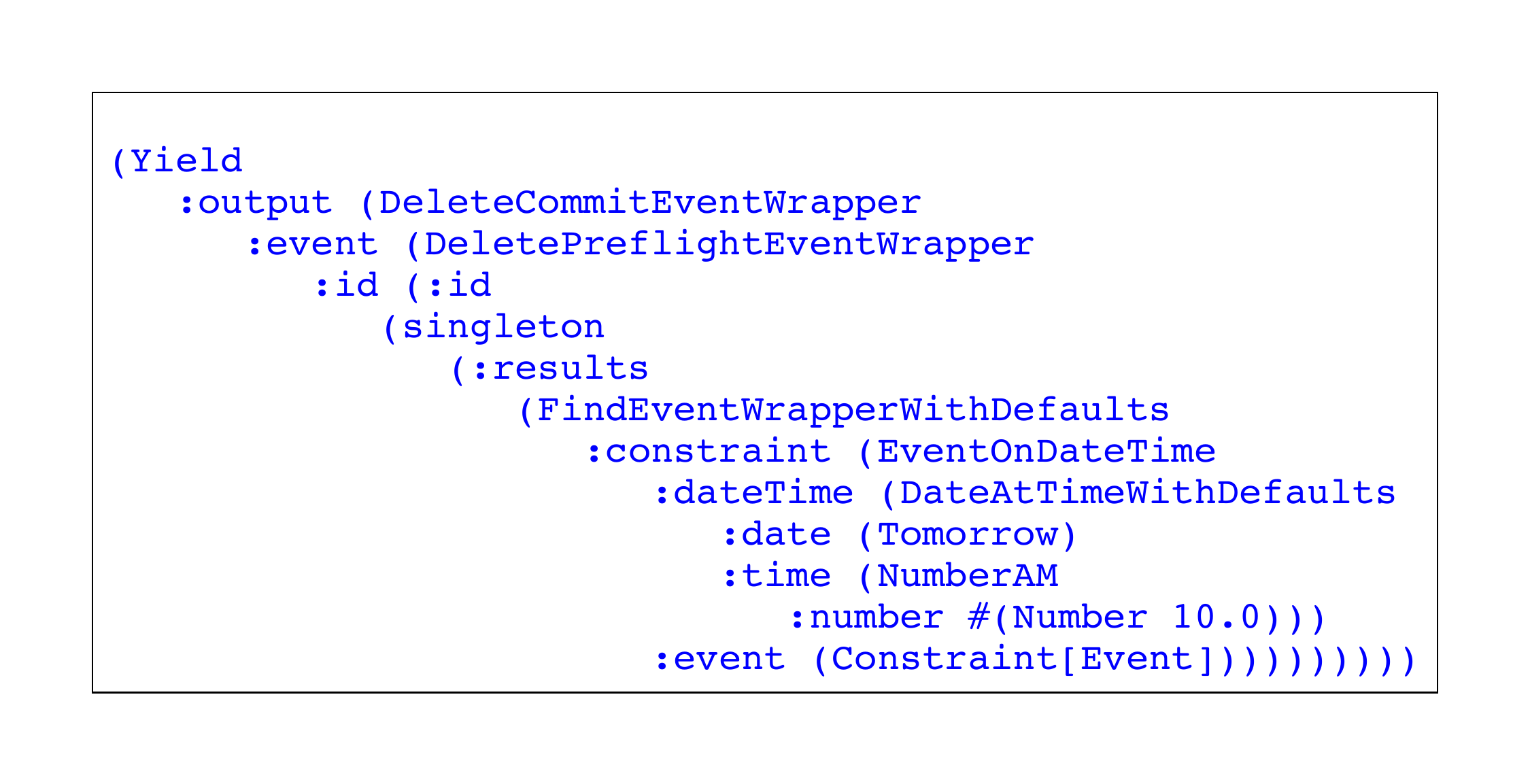}
\caption{\label{ex1-org-txt} Example 1 - Original annotation}
\end{figure}

\begin{itemize}
\item The {\it Yield} function is removed, as it is added as the top level function in {\it every} annotation. At execution time, a wrapping {\it Yield} is automatically added back. 
\item The event deletion operation is originally realized as a two step process ({\it DeletePreflightEventWrapper, DeleteCommitEventWrapper}), first ensuring the correct event is found, requesting confirmation from the user, and then deleting it through a call to an external API. In the original annotation, these steps are explicitly mentioned, but in the simplified version, this is replaced with one operation ({\it DeleteEvent}). At execution time, this is converted back to two separate functions.
\item The input to the delete event function is originally an event id (presumably a unique integer identifying the event), and the id is explicitly extracted from the found event in a separate step. In the simplified annotation, there is no need for this additional step. Instead, the implementation of {\it DeleteEvent} knows how to handle different input types. Specifically, it can apply the necessary conversions: if the input is an {\it Int}, it will be used as the event id. If the input is an {\it Event}, it will be used as is. If the input is multiple {\it Event}s, then the {\it singleton} function will be added to wrap the input. If the input is a specification of an event, then a {\it FindEvents} function is added to wrap the input.
\item The original annotation calls the function {\it DateAtTimeWithDefaults}, presumably applying some logic to fill missing time information. In the simplified annotation, this kind of logic is moved from the annotation to the function implementation.
\item The original function {\it EventOnDateTime}, with an empty "formal" parameter "{\it Constraint[Event]}" is removed in the simplification.
\item The new function {\it starts\_at} (which can handle various types of input) replaces the old functions to specify the time constraints.
\end{itemize}

The resulting simplified annotation is more concise. More importantly, it is, arguably, more understandable, and "closer" to the user request, without losing any necessary information.

\subsection{Example 2}
In this example, the user request is: 

\textsf{"Change on Sunday at Jeffs from 10:00 to 10:30 AM to 10:00 am to 2:00 pm."}

\begin{figure}
\includegraphics[height=0.4\textheight,trim= 1.2in 0.5in 0.8in 0.5in]{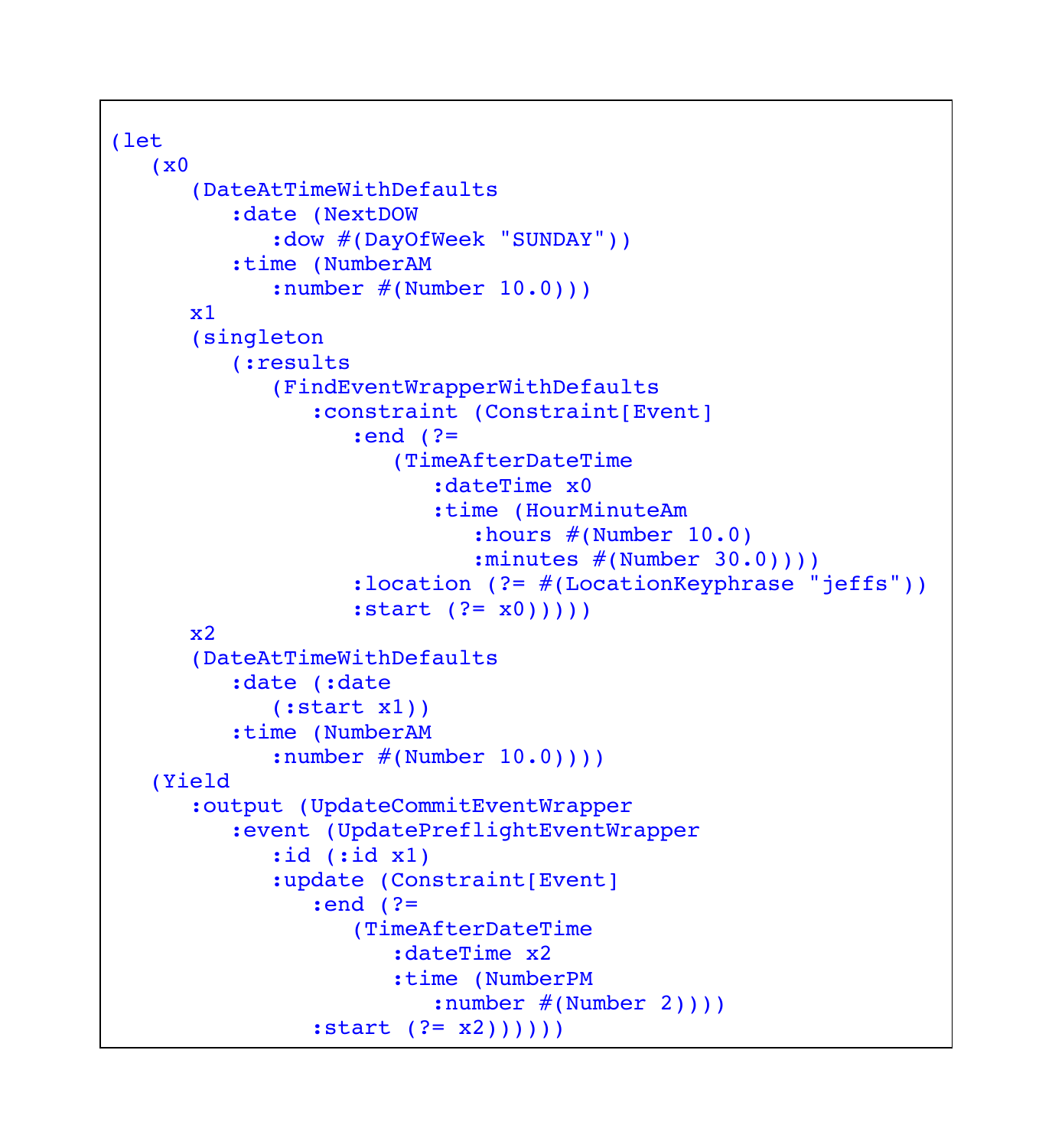}
\caption{\label{ex2-org-txt} Example 2 - Original annotation}
\end{figure}

\begin{figure}
\includegraphics[height=0.19\textheight,trim= 0.0in 0.5in 0.8in 0.5in]{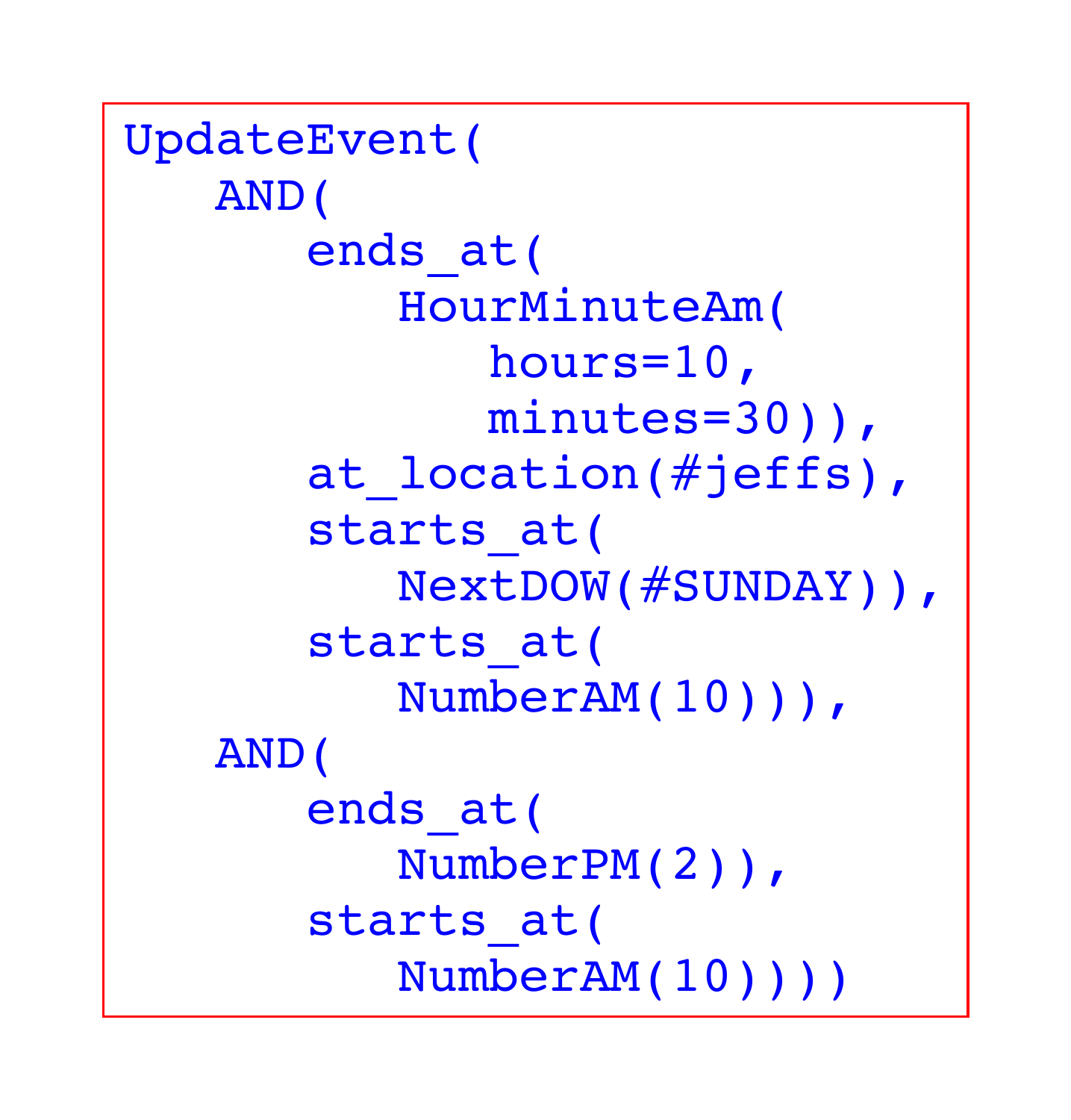}
\caption{\label{ex2-simp-txt} Example 2 - Simplified annotation}
\end{figure}

\begin{figure}
\includegraphics[height=0.3\textheight,trim= 0.99in 0.5in 0.8in 0.5in]{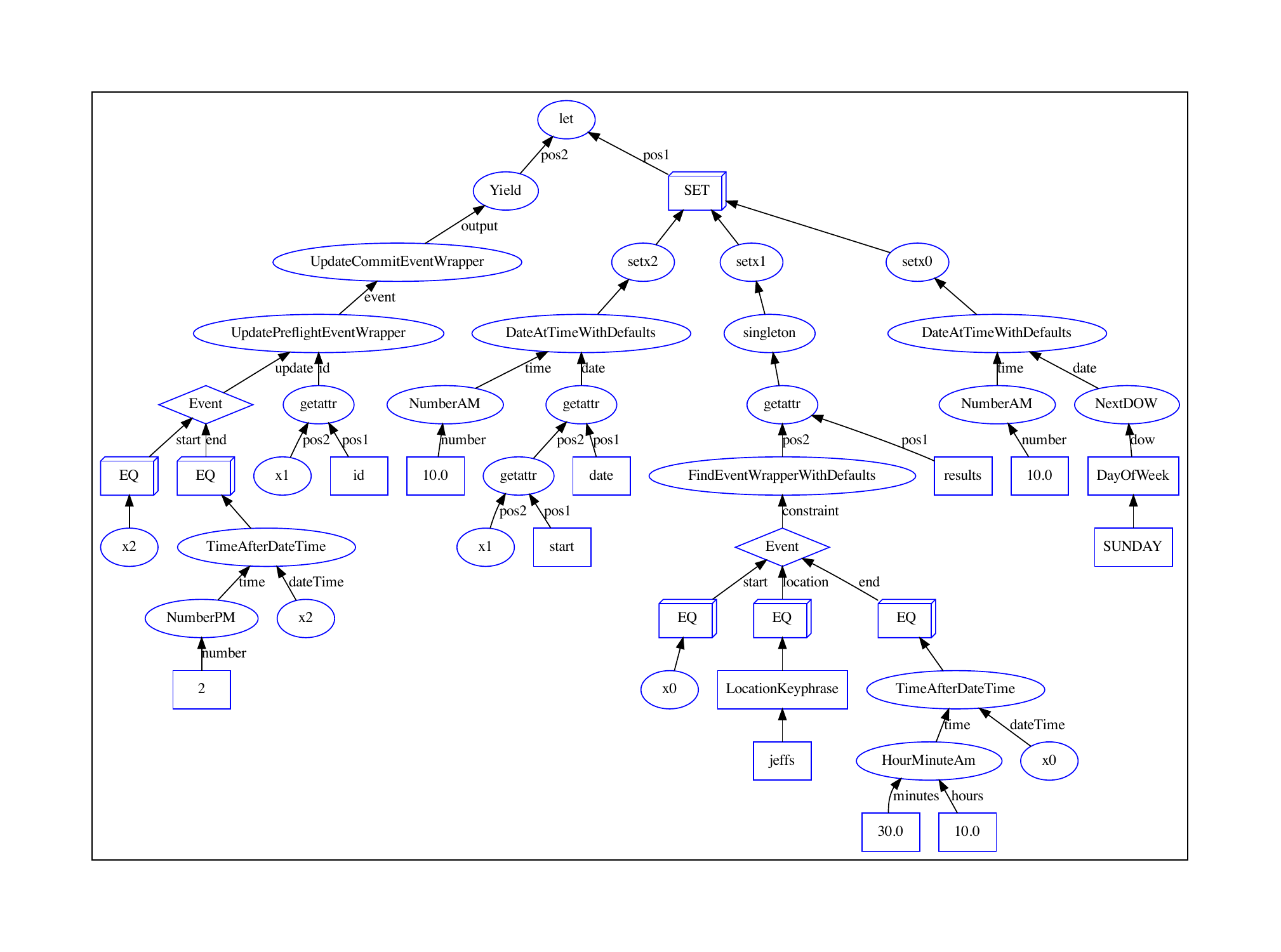}
\caption{\label{ex2-org} Example 2 - Original graph}
\end{figure}

\begin{figure}
\includegraphics[height=0.15\textheight,trim= 0.0in 0.5in 0.8in 0.5in]{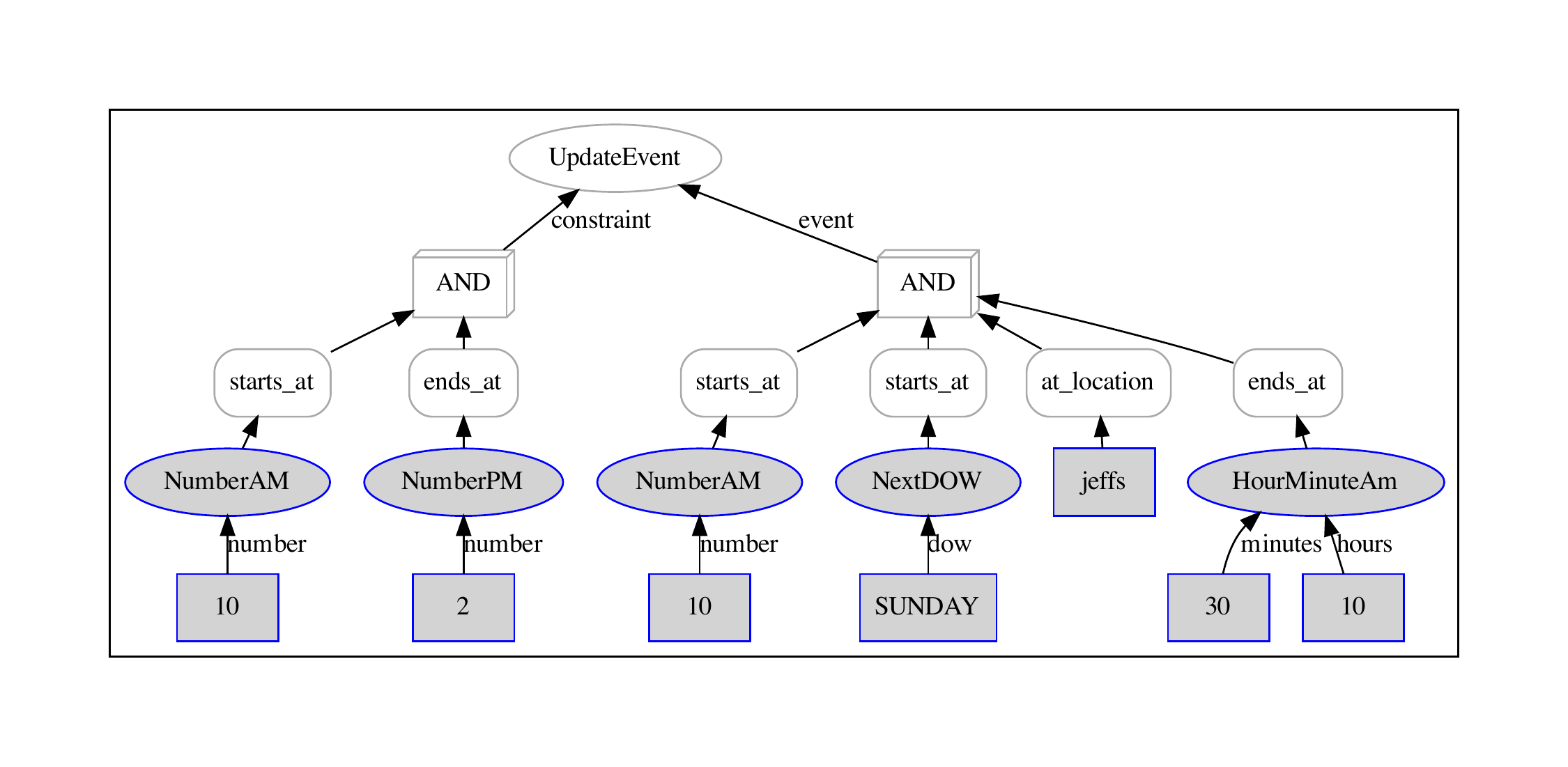}
\caption{\label{ex2-simp} Example 2 - Simplified graph}
\end{figure}

Figure \ref{ex2-org-txt} and \ref{ex2-org} show the original annotation, in text and graph formats, respectively,  for the second example. Figures \ref{ex2-simp-txt} and \ref{ex2-simp} show the same for the simplified annotation.

This example annotation uses an assignment construct, which defines separate mini-graphs which can be re-used within one annotation. While useful, it can make the annotation more difficult to understand, and require more effort to annotate. 

Assignments are kept in the simplified annotation, but specific simplification rules try to remove unnecessary use of assignments. Alternative annotation designs could altogether avoid the definition of separate mini-graphs, directly reusing parts of the main graph. 

In this example, the assignments can automatically be completely removed by the simplification rules, as after successive simplification steps, no mini-graph is used in multiple places.

Explanation of the original annotation and its simplification:
\begin{itemize}
\item Assigns label x0 to "Sunday at 10:00AM"
\item Assigns the label x1 to the computation: "find an event at location 'Jeffs', starting at x0, and ending at 10:30AM after x0". The fragment "start at x0 and end after x0" is simplified to just "start at x0". After this, x0 appears only once in the graph, so the x0 mini-graph can be directly attached to the x1 mini-graph.
\item Assigns the label x2 to "the date of x1 (the found event), and time 10AM"
\item The main computation then tries to update x1 (the found event) to start at x2 (10AM at the date of x1), and end at 2PM after x2. Updating x1 to have the date of x1 is automatically removed, leaving just "start at 10AM, end at 2PM". After this, x2 is not used any more, and x1 is used only once, so it is attached to the main computation. 
\end{itemize} 

Again, the simplified annotation is shorter and, arguably, clearer, while retaining all the necessary information of the user request.

\cleardoublepage

\cleardoublepage

\section{Example Dataflow Function}

In order to demonstrate that writing DF programs is not complex, this section shows an example implementation of a simple DF function, and using it in a non-trivial mini dialogue, taking advantage of the DF framework functionality. 

\begin{figure}[b!]
\includegraphics[height=0.3\textheight,trim= 0.43in 7in 0.8in 0.19in]{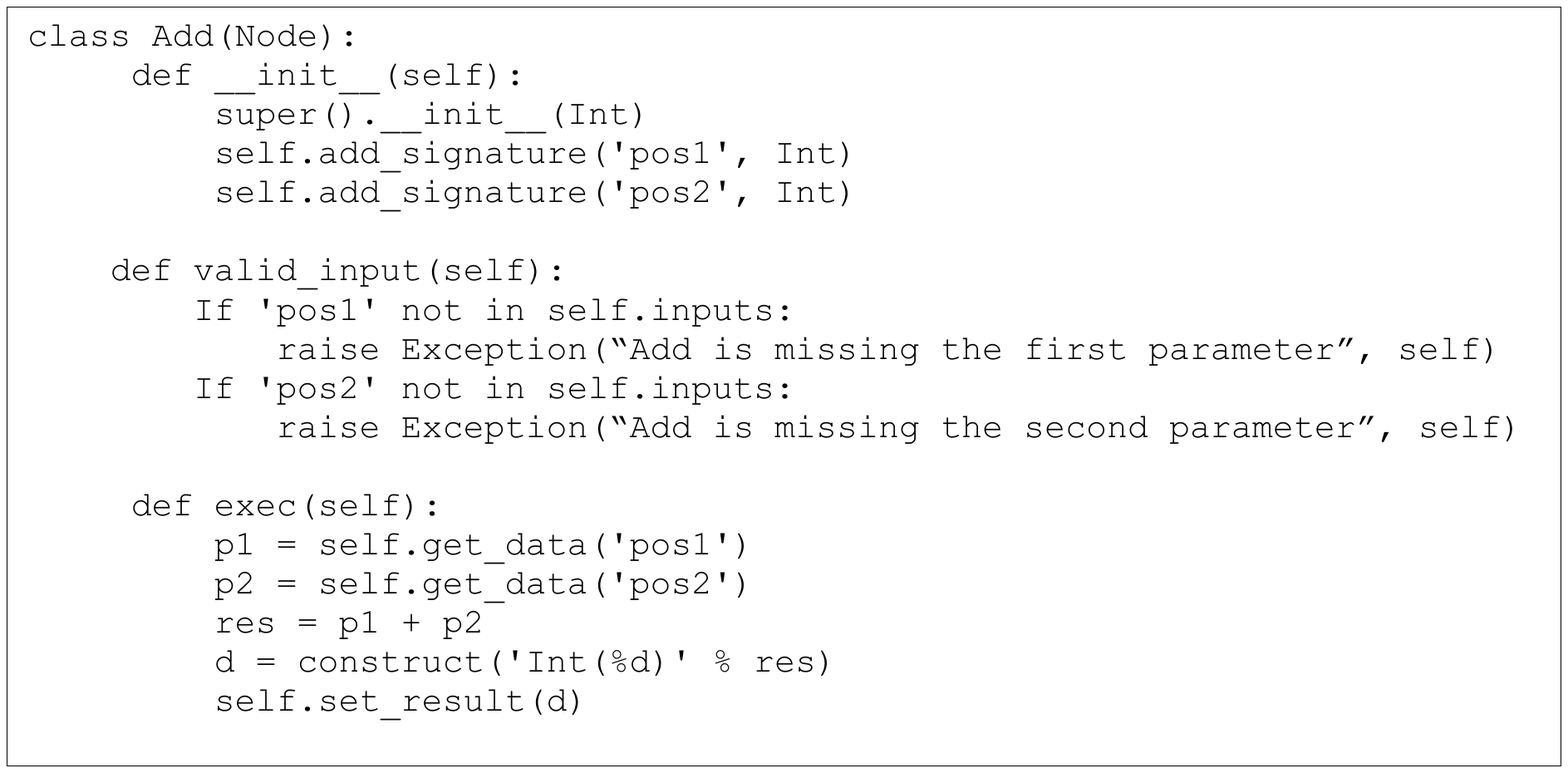}
\caption{\label{add-df} Code snippet implementing 'Add'}
\end{figure}

The code snippet implements the simple function of integer addition. The Python code in figure \ref{add-df} is (almost) executable by the released DF execution code (minor details are modified for the sake of clarity).

The definition consists of Three main blocks:
\begin{itemize}
\item Function signature definition. In this case, declaring two inputs of type {\it Int} (named {\it pos1, pos2}), and setting the output type to {\it Int}.
\item Validity checks on the inputs (optional). In this example, missing input will trigger the exception mechanism.
\item The code executing the function (optional). In this case, the values of the inputs are added, and a new {\it Int} node, with its value set to the calculated sum, is created and attached as result.
\end{itemize}

Once this function is defined, the execution engine can execute DF expressions using this function. For example, executing the expressions: 
\begin{verbatim}
'Add(2,Add(3,5))'
'revise(old=Int?(3), new=Int(6))'
\end{verbatim}
(corresponding to the user requests: {\it "add 2 to the sum of 3 and 5"}, followed by 

\begin{figure}
\includegraphics[height=0.27\textheight,trim= 0.4in 0in 0.0in 0.0in]{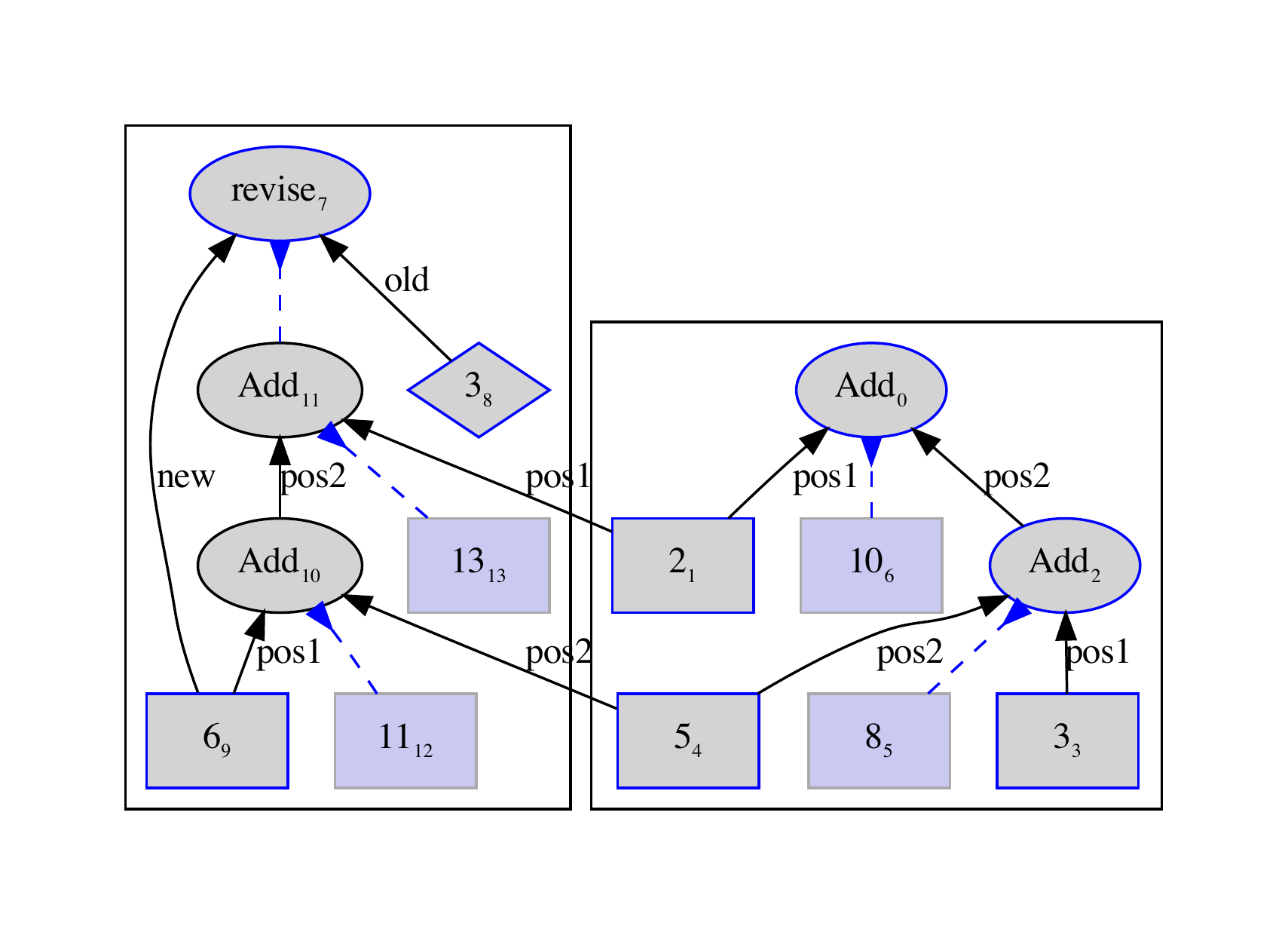}
\caption{\label{add-ex} Graph context after executing two user turns. Right: result of first turn, using the {\it Add} function (blue dashed arrows point to result nodes). Left: Result of executing the {\it revise} turn: the graph whose root is node $Add_0$ is duplicated as a graph whose root is $Add_{11}$, sharing inputs with the old graph, but replacing the original input ({\it 3}) to node $Add_2$ by the new input ({\it 6})  to $Add_{10}$ (which is the duplication of $Add_2$). }
\end{figure}

{\it "make it 6 instead of 3"}), 
builds the graphs shown in figure \ref{add-ex}.

For the first expression, it checks that inputs to all nodes are valid, and recursively executes the calculation, setting the results of all the nodes to the corresponding addition results. 

For the second expression, the system finds the relevant computational graph and the node to be replaced, and then creates a duplicate of the graph, replacing the old input node with the new one, and finally executing (evaluating) the new graph.

\end{document}